\documentclass[letterpaper]{article} 
\usepackage{aaai2026}
\usepackage{times}  
\usepackage{helvet}  
\usepackage{courier}  
\usepackage[hyphens]{url}  
\usepackage{graphicx} 
\usepackage{natbib}  
\usepackage{caption} 
\usepackage{algorithm}
\usepackage{algorithmic}

\usepackage{newfloat}
\usepackage{listings}
\DeclareCaptionStyle{ruled}{labelfont=normalfont,labelsep=colon,strut=off} 
\floatstyle{ruled}
\newfloat{listing}{tb}{lst}{}
\floatname{listing}{Listing}
\title{STMI: Segmentation-Guided Token Modulation with Cross-Modal Hypergraph Interaction for Multi-Modal Object Re-Identification}

\author {
    Xingguo Xu\textsuperscript{\rm 1}\equalcontrib,
    Zhanyu Liu\textsuperscript{\rm 1}\equalcontrib,
    Weixiang Zhou\textsuperscript{\rm 1}\equalcontrib,
    Yuansheng Gao\textsuperscript{\rm 2},\\
    Junjie Cao\textsuperscript{\rm 1}\thanks{Corresponding authors.},
    Yuhao Wang\textsuperscript{\rm 3}\footnotemark[2],
    Jixiang Luo\textsuperscript{\rm 4},
    Dell Zhang\textsuperscript{\rm 4}
}

\affiliations {
    \textsuperscript{\rm 1}School of Mathematical Sciences, Dalian University of Technology, China\\
    \textsuperscript{\rm 2}College of Computer Science and Technology, Zhejiang University, China\\
    \textsuperscript{\rm 3}School of Future Technology, Dalian University of Technology, China\\
    \textsuperscript{\rm 4}Institute of Artificial Intelligence (TeleAI), China Telecom, China\\
    xuxingguo@mail.dlut.edu.cn
}

\usepackage{bibentry}

\usepackage{algorithm}
\usepackage{algorithmic}
\usepackage{amssymb}
\usepackage{amsmath}
\usepackage{multirow}
\usepackage{booktabs}
\usepackage{pifont}
\usepackage{colortbl}
\begin{document}

\maketitle

\begin{abstract}
Multi-modal object Re-Identification (ReID) aims to exploit complementary information from different modalities to retrieve specific objects. However, existing methods often rely on hard token filtering or simple fusion strategies, which can lead to the loss of discriminative cues and increased background interference. To address these challenges, we propose \textbf{STMI}, a novel multi-modal learning framework consisting of three key components: (1) \textit{Segmentation-Guided Feature Modulation} (SFM) module leverages SAM-generated masks to enhance foreground representations and suppress background noise through learnable attention modulation; (2) \textit{Semantic Token Reallocation} (STR) module employs learnable query tokens and an adaptive reallocation mechanism to extract compact and informative representations without discarding any tokens; (3) \textit{Cross-Modal Hypergraph Interaction} (CHI) module constructs a unified hypergraph across modalities to capture high-order semantic relationships. Extensive experiments on public benchmarks (i.e., RGBNT201, RGBNT100, and MSVR310) demonstrate the effectiveness and robustness of our proposed STMI framework in multi-modal ReID scenarios.

\end{abstract}
\begin{figure}[t]
  \centering
    \resizebox{0.43\textwidth}{!}
    {
  \includegraphics[width=30\linewidth]{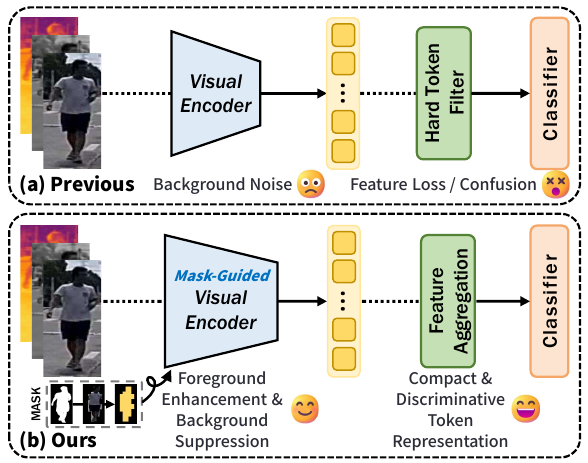}
  }
   \caption{Motivation and intuitive comparison. (a) Existing methods suffer from background noise and information loss due to hard token filtering. (b) Our proposed STMI framework introduces segmentation-guided modulation module to enhance foreground and suppress background, enabling more discriminative feature learning across modalities.}
  \label{fig:chuangxin}
  \vspace{-2mm}
\end{figure}
\section{Introduction}

In recent years, multi-modal object Re-Identification (ReID) has attracted increasing attention due to its wide range of applications in practical scenarios such as intelligent surveillance, cross-spectrum monitoring, and nighttime recognition. Unlike traditional RGB-based ReID~\cite{liu2021watching,zhang2021hat,wang2021pyramid,shi2024learning}, multi-modal object ReID involves multiple visual modalities, including visible light (RGB), near-infrared (NIR), and thermal infrared (TIR), offering enhanced robustness under challenging conditions such as drastic illumination changes, low-light environments, or nighttime scenes~\cite{zhao2023multimodal,he2023low,zheng2025collaborative,TangLZHT25}. However, due to the significant distribution discrepancies across different modalities, achieving effective multi-modal representation learning remains a fundamental challenge.

Existing methods primarily focus on aligning and fusing visual features, typically leveraging strategies such as token selection, modality transformation, or attention mechanisms to process multi-modal images~\cite{wan2025ugg,li2025icpl,wan2025reliable,lin2025dmpt,bian2025modality}, as illustrated in Fig.~\ref{fig:chuangxin}. However, these approaches suffer from two major limitations. First, during token selection, “redundant” regions are often removed via hard cropping, which may inadvertently discard critical details and compromise discriminative performance. Second, in multi-modal feature fusion, the lack of effective modeling of high-order semantic relationships limits the ability to fully exploit complementary information across modalities, especially in complex scenes with background clutter or occlusions.

To address the aforementioned issues, we propose a feature learning framework named \textbf{STMI}, which introduces \textbf{S}egmentation-guided \textbf{T}oken \textbf{M}odulation with cross-modal hypergraph \textbf{I}nteraction for multi-modal object ReID. First, we introduce the Segmentation-Guided Feature Modulation (SFM) module, which leverages foreground masks generated by the SAM segmentation model to guide attention learning. Specifically, we incorporate two learnable modulation parameters to adaptively reweight token features, emphasizing foreground regions while suppressing background noise. Second, we propose the Semantic Token Reallocation (STR) module to refine the token representation in a more structured manner. Rather than relying on hard filtering strategies, STR introduces multiple learnable query tokens that interact with patch tokens via a cross-attention mechanism. This enables the extraction of compact, informative semantic representations while preserving fine-grained visual details. Third, we design the Cross-Modal Hypergraph Interaction (CHI) module to capture high-order semantic relationships across different modalities. In this module, semantic tokens from RGB, NIR, and TIR images are treated as nodes within a unified hypergraph. Cross-modal hyperedges are constructed based on semantic similarity, allowing the model to learn structural correlations among local regions across modalities. By jointly leveraging segmentation priors, semantic token reconstruction, and high-order relational modeling, STMI effectively maintains token completeness and enhances feature discrimination, achieving superior performance in challenging multi-modal ReID scenarios.
Our main contributions are summarized as follows:
\begin{itemize}
    \item We propose STMI, a novel multi-modal ReID framework. To the best of our knowledge, it is the first work to incorporate segmentation masks for attention modulation in multi-modal object ReID.
    
    \item We introduce a Segmentation-Guided Feature Modulation (SFM) module that enhances foreground regions and suppresses background interference, preserving discriminative information without discarding any tokens.
    
    \item We design a Semantic Token Reallocation (STR) module based on cross-attention, which extracts structured and compact semantic tokens using learnable queries, avoiding information loss caused by hard token filtering.
    
    \item We present a Cross-Modal Hypergraph Interaction (CHI) module that models high-order semantic relationships across modalities by constructing a unified hypergraph, enabling rich inter-modal dependency modeling.
    
    \item Extensive experiments on three public multi-modal ReID datasets demonstrate that STMI achieves state-of-the-art performance, validating its effectiveness and robustness.
\end{itemize}

\section{Related Work}

\subsection{Multi-Modal Object Re-Identification}

Benefiting from the complementary information across modalities, multi-modal ReID has demonstrated superior stability and performance. Existing approaches primarily focus on modeling cross-modal interactions~\cite{zhang2025prompt,yang2025tienet,fengmulti}. For instance, TOP-ReID~\cite{wang2024top} introduces a cyclic interaction mechanism via cross-attention to fuse tri-modal features. MambaPro~\cite{wang2024mambapro} adopts the Mamba architecture~\cite{gu2023mamba} to capture both intra-modal and inter-modal dependencies. Furthermore, DeMo~\cite{wang2024decoupled} proposes an adaptive Mixture of Experts (MoE) framework to decouple modality-specific information and perform weighted cross-modal feature aggregation. However, most of these methods model all tokens across the entire image, making them vulnerable to background noise, which deteriorates feature quality and limits overall performance~\cite{tian2018eliminating}. To address this issue and better focus on informative regions, several works explore key feature extraction prior to cross-modal fusion. EDITOR~\cite{zhang2024magic} leverages attention maps to select salient features, guiding the model to attend to important regions. IDEA~\cite{wang2025idea} samples tokens from key spatial locations and adaptively learns positional shifts to capture fine-grained local details. NEXT~\cite{li2025next} introduces textual cues to guide context-aware token sampling. While these sampling-based strategies help the model focus on salient regions, they also introduce new challenges. The adaptiveness of token selection does not always ensure the most informative features. This may result in \textbf{semantic loss} and \textbf{feature confusion} with hard token pruning, ultimately hindering performance. To overcome this limitation, we propose a token modulation strategy that \textbf{preserves critical information} more effectively, while also \textbf{mitigating noise and ambiguity} in the feature representation.

\subsection{Semantic Segmentation for Feature Enhancement}
Semantic segmentation has seen significant advancements in various visual tasks, particularly in handling complex image editing and instance segmentation, where it demonstrates powerful capabilities. In recent years, pre-trained models such as OpenPifPaf~\cite{kreiss2021openpifpaf}, SAM~\cite{kirillov2023segany}, and SAM2~\cite{ravi2024sam2} have demonstrated strong generalization capabilities across various vision scenarios. These models can generate masks containing rich semantic information, which can be effectively utilized in a wide range of visual tasks. For instance, VideoGrain~\cite{yang2025videograin} leverages SAM for instance segmentation in multi-grained video editing tasks, generating precise masks to help control the editing of different parts of the video. Additionally, VoteSplat~\cite{jiang2025votesplat} integrates SAM with a Hough voting mechanism to achieve accurate instance segmentation. SmartFreeEdit~\cite{sun2025smartfreeedit} uses SAM to generate reasoning segmentation masks, supporting image editing guided by natural language instructions. While in ReID tasks, for instance, Mask-Guided~\cite{song2018mask} introduces binary masks to guide the generation of body and background attention maps for region-level learning. MaskReID~\cite{qi2019mask} directly uses segmentation results along with RGB images for improved feature representations. Nonetheless, prior works incorporate segmentation masks merely as auxiliary inputs, lacking \textbf{fine-grained} and \textbf{token-level modulation}. In contrast, our proposed STMI framework embeds segmentation into the attention mechanism, enabling more precise and consistent feature enhancement across modalities.

\begin{figure}[t]
    \centering
      \resizebox{0.48\textwidth}{!}
      {
    \includegraphics[width=30\linewidth]{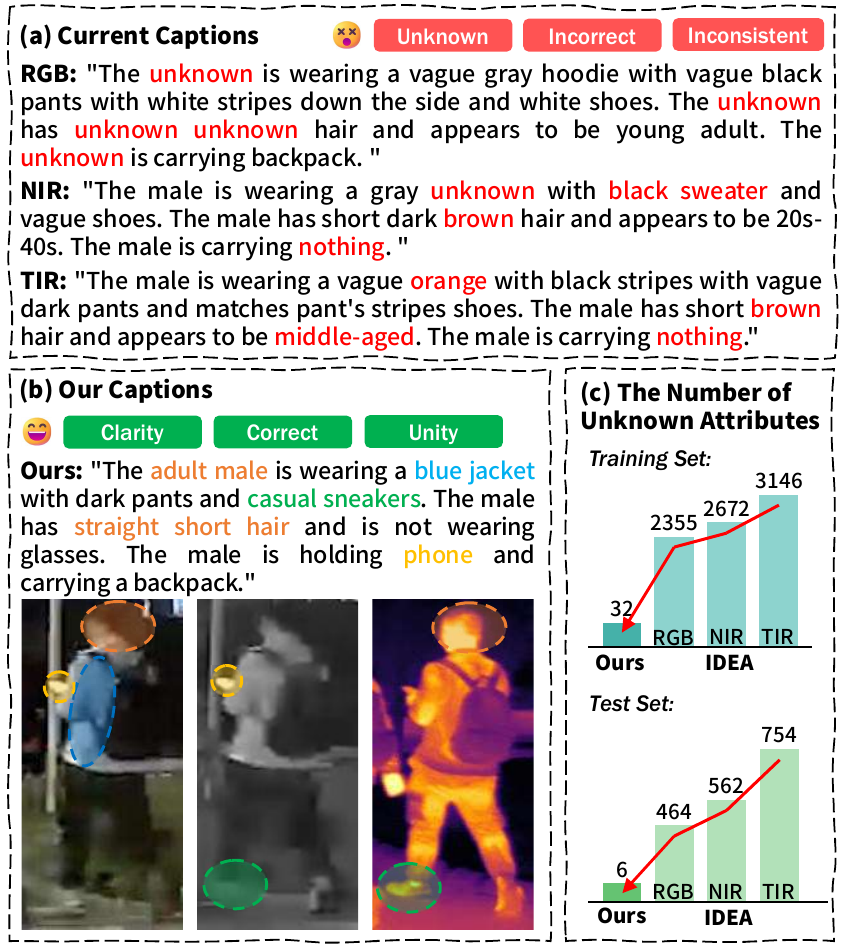}
    }
        \vspace{-4mm}
     \caption{Comparison with IDEA: (a) IDEA captions often include unknown or inconsistent attributes; (b) ours generates clearer and more accurate descriptions across modalities; (c) our method significantly reduces unknown attributes in both training and test sets.}
    \label{fig:text}
    \vspace{-4mm}
  \end{figure}
  
\section{Proposed Method}

\begin{figure*}[t]
    \centering
      \resizebox{1.0\textwidth}{!}
      {
    \includegraphics[width=30\linewidth]{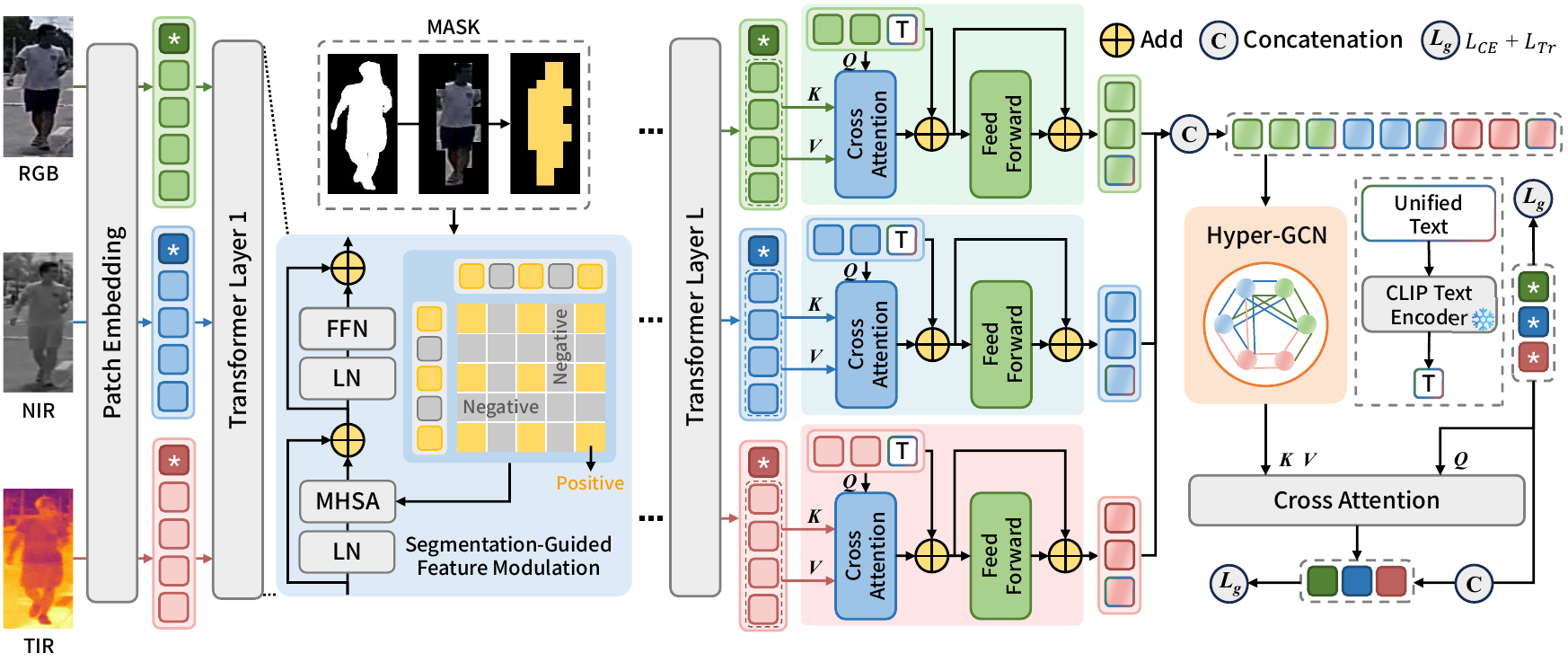}
    }
    \vspace{-5mm}
     \caption{An overview of our proposed STMI framework, which consists of three main modules: (1) Segmentation-Guided Feature Modulation enhances foreground and suppresses background using SAM masks; (2) Semantic Token Reallocation extracts compact semantic tokens via cross-attention with learnable queries; (3) Cross-Modal Hypergraph Interaction builds a hypergraph across modalities for high-order semantic interaction.}
    \label{fig:Overall}
    \vspace{-4mm}
  \end{figure*}

As shown in Fig.~\ref{fig:Overall}, our proposed \textbf{STMI} consists of three components: Segmentation-Guided Feature Modulation (SFM) module, Semantic Token Reallocation (STR) module, and Cross-Modal Hypergraph Interaction (CHI) module. Below, we describe each component in detail.

\subsection{Multi-Modal Caption Generation}

In multi-modal object ReID tasks, introducing semantic descriptions as auxiliary guidance~\cite{wang2025idea} has been shown to significantly improve model performance. However, existing text generation methods still suffer from several major limitations, as illustrated in Fig.~\ref{fig:text}: (1) \textbf{Modality inconsistency:} Most existing approaches generate textual descriptions based solely on a single modality (e.g., RGB), ignoring complementary semantic cues potentially present in other modalities such as NIR or TIR. This often leads to incomplete or biased descriptions; (2) \textbf{Semantic ambiguity:} Under challenging conditions such as occlusion, low light, or blur, multi-modal large language models (MLLMs) often fail to identify key attributes, resulting in vague responses like ``unknown'' or even refusal to answer. (3) \textbf{Lack of confidence estimation:} Most existing methods do not provide confidence scores for each generated attribute, making it difficult to assess the reliability of the semantic information.

To address the above issues, we propose two strategies to enhance the quality and reliability of multi-modal caption generation. First, we adopt an image concatenation-based input, where images of the same identity from three modalities are concatenated into a single composite image and fed into an MLLM. This design enables the model to perceive multi-modal information holistically and generate more complete and consistent natural language descriptions. Second, we introduce a structured attribute extraction and confidence-aware filling strategy, inspired by NEXT~\cite{li2025next}, which leverages attribute-level confidence to guide text generation. Specifically, we first use the MLLM to extract attribute–value–confidence triplets from each individual modality as well as the concatenated multi-modal image. These triplets are then fed into an LLM along with a predefined template, prompting it to select the most reliable attribute values based on confidence scores and generate the final description. \textbf{As shown in Fig.~\ref{fig:text}}, these two strategies significantly enhance the reliability and consistency of the generated textual descriptions, thereby providing high-quality semantic information for downstream multi-modal feature modeling.

\subsection{Segmentation-Guided Feature Modulation}

To enhance the model's ability to focus on foreground regions while suppressing background interference, we propose the \textit{Segmentation-Guided Feature Modulation} (SFM) module. This module explicitly guides the attention maps in self-attention layers using semantic segmentation masks, enabling region-aware feature modeling. Specifically, given an input image $I \in \mathbb{R}^{3 \times H \times W}$ and its corresponding binary segmentation mask $M \in \{0,1\}^{H \times W}$, we first divide the image into $N$ patches through a vision encoder and extract $N+1$ token representations, including one class token as follows:
\begin{equation}
    F = [f_{\text{cls}}, f_1, f_2, \dots, f_N] \in \mathbb{R}^{(N+1) \times D},
\end{equation}
where ${f}_{\text{cls}}$ denotes the class token, and the rest are patch tokens, with $D$ being the feature dimension per token.
Next, we construct a token-level binary mask based on the spatial overlap between each patch and the segmentation mask:
\begin{equation}
    {m} = [1, m_1, m_2, \dots, m_N] \in \{0,1\}^{N+1},
\end{equation}
where the first position is set to 1, treating the class token as part of the foreground. The remaining $m_i$ indicate whether the $i$-th patch token lies within the foreground region.

At the $l$-th Transformer layer, the self-attention module~\cite{vaswani2017attention} first computes the attention logits:
\begin{equation}
    {A}_{\text{logit}}^{(l)} = {Q}^{(l)} ({K}^{(l)})^\top,
\end{equation}
where ${Q}^{(l)}, {K}^{(l)} \in \mathbb{R}^{(N+1) \times D}$, and $D$ is the attention dimension. We then construct positive and negative modulation matrices based on ${A}_{\text{logit}}^{(l)}$, used for enhancing foreground and suppressing background regions, respectively:
\begin{align}
    {M}^{\text{pos}} &= \max({A}_{\text{logit}}^{(l)}) - {A}_{\text{logit}}^{(l)}, \\
    {M}^{\text{neg}} &= {A}_{\text{logit}}^{(l)} - \min({A}_{\text{logit}}^{(l)}),
\end{align}
where $\max(\cdot)$ and $\min(\cdot)$ are row-wise operations, broadcasting the maximum/minimum value of each row across the entire row. The resulting matrices ${M}^{\text{pos}}, {M}^{\text{neg}} \in \mathbb{R}^{(N+1)\times(N+1)}$ represent the relative deviation of each token pair. To guide the attention modulation, we construct a foreground interaction mask ${R}$ based on ${m}$ as follows:
\begin{equation}
    {R} = {m}^\top \cdot {m} \in \{0,1\}^{(N+1)\times(N+1)},
\end{equation}

where ${R}[i,j] = 1$ if both tokens $i$ and $j$ are foreground. Otherwise, ${R}[i,j] = 0$, treating the pair as background.

Considering that the semantic segmentation mask may suffer from boundary ambiguity or incorrect segmentation, we introduce a mask perturbation mechanism during training to enhance model robustness. Specifically, only background tokens are perturbed with a probability $p$ by flipping their label to foreground (i.e., $m_i := 1$ if $m_i = 0$), while foreground tokens remain unchanged. This strategy effectively mitigates the overfitting risk caused by mask guidance and improves the generalization ability of the model. Thus, the final modulation matrix ${S}$ is defined as:
\begin{equation}
    {S} = \alpha \cdot {R} \odot {M}^{\text{pos}} - \beta \cdot (1 - {R}) \odot {M}^{\text{neg}},
\end{equation}
where $\odot$ denotes element-wise multiplication, and $\alpha$, $\beta$ are learnable parameters that control the strength of foreground enhancement and background suppression. Then the modulated attention weights are computed as follows:
\begin{equation}
    \hat{{A}}^{(l)} = \text{softmax}\left( \frac{{A}_{\text{logit}}^{(l)} + {S}}{\sqrt{D}} \right).
\end{equation}
Through the above mechanism, SFM enhances the model’s semantic understanding of foreground regions while effectively suppressing background noise during multi-modal feature modeling in the backbone.

\subsection{Semantic Token Reallocation}
To enhance semantic alignment across modalities, we propose the \textit{Semantic Token Reallocation} (STR) module. It employs learnable, modality-specific query tokens, along with a shared global text feature, to guide the cross-attention process. This design enables a structured reconstruction of visual semantic tokens, leading to improved cross-modal consistency. Specifically, for each modality $m \in \{\text{RGB}, \text{NIR}, \text{TIR}\}$, the input is a sequence of patch tokens extracted from the backbone:
\begin{equation}
    {F}^{(m)} = \left[{f}^{(m)}_1, {f}^{(m)}_2, \dots, {f}^{(m)}_N \right] \in \mathbb{R}^{N \times D}.
\end{equation}
For each modality, we introduce $K$ independent learnable semantic query tokens, with the following equation:
\begin{equation}
{Q}^{(m)} = \left[{q}^{(m)}_1, \dots, {q}^{(m)}_K \right] \in \mathbb{R}^{K \times D}.
\end{equation}
To incorporate a cross-modal semantic prior, we further extract a shared global textual feature using the text encoder of CLIP~\cite{radford2021learning}, which captures the overall description of the image across all three modalities. This shared global feature is denoted as ${T} \in \mathbb{R}^{D}$. We concatenate it to the end of the semantic query token sequence to form an enhanced query sequence:
\begin{equation}
{Q}'^{(m)} = \left[{Q}^{(m)}; {T} \right] \in \mathbb{R}^{(K+1) \times D}.
\end{equation}
Here, ${Q}'^{(m)}$ serves as the query, while ${F}^{(m)}$ serves as both key and value. A cross-attention operation followed by a Feed-Forward Network (FFN)~\cite{vaswani2017attention} is applied to obtain the final semantic token representations:
\begin{equation}
{Z}^{(m)} = \text{CrossAttn}({Q}'^{(m)}, {F}^{(m)}, {F}^{(m)}) + {Q}'^{(m)},
\label{eq:crossattn_output}
\end{equation}
\begin{equation}
\tilde{{F}}^{(m)} = \text{FFN}({Z}^{(m)}) + {Z}^{(m)}.
\label{eq:ffn_output}
\end{equation}
The resulting $\tilde{{F}}^{(m)}$ is then used as input to the multi-modal semantic alignment and interaction modeling stage, serving as the basis for subsequent cross-modal feature fusion.

\subsection{Cross-Modal Hypergraph Interaction}

After obtaining the semantic token representations for each modality, we design the \textit{Cross-Modal Hypergraph Interaction} (CHI) module to model high-order semantic relationships across modalities. Unlike traditional graph structures, hypergraphs can connect multiple nodes within a single hyperedge, making them naturally suitable for joint modeling of multi-modal information. Specifically, let the semantic tokens from the three modalities be denoted as $\tilde{{F}}^{(\text{RGB})}$, $\tilde{{F}}^{(\text{NIR})}$, and $\tilde{{F}}^{(\text{TIR})}$, each of shape $\mathbb{R}^{(K+1) \times D}$. We first concatenate them into a unified cross-modal semantic token set:
\begin{equation}
{H} = \left[ \tilde{{F}}^{(\text{RGB})}; \tilde{{F}}^{(\text{NIR})}; \tilde{{F}}^{(\text{TIR})} \right] \in \mathbb{R}^{3(K+1) \times D}.
\label{eq:concat_modal_features}
\end{equation}
Based on this, we construct a cross-modal hypergraph $\mathcal{G} = (\mathcal{V}, \mathcal{E})$, where the node set $\mathcal{V}$ corresponds to each token in ${H}$, i.e., $|\mathcal{V}| = 3(K+1)$. The hyperedge set $\mathcal{E}$ is dynamically generated based on intra- and inter-modal semantic similarities. Specifically, we compute a similarity matrix ${S} \in \mathbb{R}^{3(K+1) \times 3(K+1)}$ over all nodes. Two nodes $i$ and $j$ are connected within the same hyperedge if their similarity $s_{ij} \geq \tau$. This allows each hyperedge to capture high-order associations and enables effective cross-modal semantic propagation. To further model information flow within the hypergraph, we introduce a hypergraph convolution operation~\cite{bai2021hypergraph}. For the node features at the $l$-th layer ${H}^{(l)} \in \mathbb{R}^{3(K+1) \times D}$, their update process in the hypergraph is formulated as:
\begin{equation}
h_i^{(l+1)} = \sigma\left( \sum_{e \in \varepsilon(i)} w_e h_e^{(l)} + b_i \right).
\end{equation}
Here, $h_e^{(l)}$ denotes the feature of hyperedge $e$ at the $l$-th iteration.
The set $\varepsilon(i)$ contains all hyperedges incident to node $i$.
Each hyperedge $e$ is associated with a learnable $w_e$.
The term $b_i$ is a node-specific bias.
$\sigma(\cdot)$ is an activation function. Specifically, we adopt a node-to-hyperedge and hyperedge-to-node mechanism, where features are first aggregated from nodes to hyperedges, and then redistributed back to nodes.
This operation allows each node to aggregate information from a group of connected nodes via hyperedges, thereby enhancing inter-modal interaction and fusion. Additionally, to preserve the independent semantic information of each original modality, we introduce a residual connection:
\begin{equation}
{H}^{(l+1)} = {H}^{(l+1)} + {H}^{(l)}.
\end{equation}
Through the hypergraph convolution, the CHI module enables high-order semantic interactions across modalities, capturing complex and rich semantic dependencies between them, thus improving the final fused representation. The resulting multi-modal semantic tokens are denoted as ${H}' \in \mathbb{R}^{3(K+1) \times D}$.

Although the CHI module effectively models high-order semantic relationships across modalities, its output semantic tokens mainly focus on local regions and still lack explicit alignment with global semantic concepts. We further employ a cross-attention mechanism, using global image features as queries to selectively aggregate complementary information from the multi-modal semantic tokens. Specifically, we extract image-level global features from the three modalities, denoted as ${g}^{\text{RGB}}$, ${g}^{\text{NIR}}$, and ${g}^{\text{TIR}}$, respectively. 

Then, we concatenate these global features into a query, as follows:
\begin{equation}
{G} = \left[{g}^{\text{RGB}}; {g}^{\text{NIR}}; {g}^{\text{TIR}} \right] \in \mathbb{R}^{3 \times D}.
\label{eq:modal_tokens}
\end{equation}
Next, we use the fused semantic token representation ${H}'$ from the CHI module as key and value pairs to construct a cross-modal cross-attention algorithm:
\begin{equation}
{U} = \text{CrossAttn}({G}, {H}', {H}') + {G},
\end{equation}
where ${U}$ represents the final fused global features. This process enables selective extraction of information relevant to global concepts from the multi-modal semantic tokens and feeds it back into the global representation.

\subsection{Objective Function}

We apply supervision to key representations, including the concatenated global feature ${G}$, fused semantic feature ${U}$, and global text feature ${T}$. For each feature representation $\mathcal{F}$, we jointly employ a label-smoothed cross-entropy loss~\cite{szegedy2016rethinking} and a triplet loss~\cite{hermans2017defense}, defined as follows:
\begin{equation}
\mathcal{L}_{g}(\mathcal{F}) = \mathcal{L}_{\text{CE}}(\mathcal{F}) + \mathcal{L}_{\text{Tri}}(\mathcal{F}),
\end{equation}
where $\mathcal{L}_{\text{CE}}$ denotes the label-smoothed cross-entropy loss, and $\mathcal{L}_{\text{Tri}}$ represents the triplet loss. The final overall loss function is defined as:
\begin{equation}
\mathcal{L} = \mathcal{L}_{g}({G}) + \mathcal{L}_{g}({U}) + \mathcal{L}_{g}({T}).
\end{equation}

\begin{table}[t]
  \centering
  \resizebox{0.475\textwidth}{!}
{
  \begin{tabular}{ccccc}
      \noalign{\hrule height 1pt}
  \multicolumn{1}{c}{\multirow{2}{*}{\textbf{Methods}}} & \multicolumn{4}{c}{\textbf{RGBNT201}} \\ \cmidrule(r){2-5}
  & \textbf{mAP} & \textbf{R-1} & \textbf{R-5} & \textbf{R-10} \\ \hline
  PFNet~\cite{zheng2021robust}    & 38.5         & 38.9            & 52.0            & 58.4             \\
  IEEE~\cite{wang2022interact}     & 47.5         & 44.4            & 57.1            & 63.6             \\
  DENet~\cite{zheng2023dynamic}    & 42.4         & 42.2            & 55.3            & 64.5            \\
  LRMM~\cite{wu2025lrmm} & 52.3 & 53.4 & 64.6 & 73.2\\
  UniCat$^*$~\cite{crawford2023unicat}   & 57.0         & 55.7            & -            & -            \\
  HTT$^*$~\cite{wang2024heterogeneous} &71.1 &73.4 &83.1 &87.3\\
  TOP-ReID$^*$~\cite{wang2024top}  &72.3 &76.6 &84.7 &89.4\\
  EDITOR$^*$~\cite{zhang2024magic} & 66.5       & 68.3           & 81.1        & 88.2             \\
  RSCNet$^*$~\cite{yu2024representation} & 68.2 & 72.5 & - & - \\
  WTSF-ReID$^*$~\cite{yu2025wtsf} & 67.9 &72.2 &83.4 &89.7 \\
  MambaPro$^\dagger$~\cite{wang2024mambapro} & 78.9 & \underline{83.4} & 89.8 & 91.9 \\
  DeMo$^\dagger$~\cite{wang2024decoupled}  &79.0 	 & 82.3	 &88.8 	 &\underline{92.0}      \\
  IDEA$^\dagger$~\cite{wang2025idea} &\underline{80.2} 	 &82.1 	 & \underline{90.0} 	 &\textbf{93.3}      \\
  \rowcolor[gray]{0.92}  
  $\mathrm{\textbf{STMI}}^\dagger$  &\textbf{81.2} 	 &\textbf{83.4} 	 &\textbf{90.2} 	 & 91.6      \\
  \noalign{\hrule height 1pt}
  \end{tabular}
  }
  \caption{Performance comparison on RGBNT201. 
  Best results are in bold, the second bests are underlined. 
  $\dagger$ denotes CLIP-based methods, $*$ indicates ViT-based while others are CNN-based ones.}
  \label{tab:multi-spectral person ReID}
\end{table}

\section{Experiments}

\subsection{Datasets and Evaluation Protocols}

\textbf{Dataset Setup.} To evaluate the effectiveness of the proposed method in complex multi-modal scenarios, we conduct experiments on three public multi-modal object ReID datasets. To improve annotation efficiency, we utilize the GPT-4o model~\cite{hurst2024gpt} provided by OpenAI to automatically generate one textual description for each image triplet. In addition, we employ the SAM2~\cite{ravi2024sam2} to generate one high-quality segmentation mask per triplet. Specifically, \textit{RGBNT201}~\cite{zheng2021robust} contains 4,787 triplets and 4,787 textual descriptions, with an average length of 33.28 characters, covering 13 semantic attributes. The \textit{MSVR310}~\cite{zheng2023cross} dataset consists of 2,087 triplets and corresponding descriptions, each with an average length of 31.51 characters, covering 6 attributes. \textit{RGBNT100}~\cite{li2020multi} is the largest dataset among them, containing 17,250 triplets and 17,250 textual annotations with an average description length of 31.90 characters, also covering 6 semantic attributes. 

\noindent \textbf{Evaluation Protocol.} We adopt mean Average Precision (mAP) and Cumulative Matching Characteristics (CMC) at ranks 1, 5, and 10 as evaluation metrics.

\begin{table}[t]
    \centering
    \resizebox{0.475\textwidth}{!}
    {
    \begin{tabular}{ccccc}
        \noalign{\hrule height 1pt}
        \multicolumn{1}{c}{\multirow{2}{*}{\textbf{Methods}}} &  \multicolumn{2}{c}{\textbf{RGBNT100}} & \multicolumn{2}{c}{\textbf{MSVR310}} \\\cmidrule(r){2-3} \cmidrule(r){4-5}
        & \textbf{mAP} & \textbf{R-1} & \textbf{mAP} & \textbf{R-1} \\
        \hline
        PFNet~\cite{zheng2021robust}& 68.1 & 94.1 & 23.5 & 37.4 \\
        GAFNet~\cite{guo2022generative} & 74.4 & 93.4 & - & - \\
        GPFNet~\cite{he2023graph} & 75.0 & 94.5 & - & - \\
        CCNet~\cite{zheng2023cross} & 77.2 & 96.3 & 36.4 & 55.2 \\
        LRMM~\cite{wu2025lrmm} & 78.6 & 96.7 & 36.7 & 49.7 \\
        GraFT$^*$~\cite{yin2023graft} & 76.6 & 94.3 & - & - \\
        UniCat$^*$~\cite{crawford2023unicat} & 79.4 & 96.2 & - & - \\
        PHT$^*$~\cite{pan2023progressively} & 79.9 & 92.7 & - & - \\
        HTT$^*$~\cite{wang2024heterogeneous} & 75.7 & 92.6 & - & - \\
        TOP-ReID$^*$~\cite{wang2024top} & 81.2 & 96.4 & 35.9 & 44.6 \\
        EDITOR$^*$~\cite{zhang2024magic} & 82.1 & 96.4 & 39.0 & 49.3 \\
        FACENet$^*$~\cite{zheng2025flare} & 81.5 & 96.9 & 36.2 & 54.1 \\
        RSCNet$^*$~\cite{yu2024representation} & 82.3 & 96.6 & 39.5 & 49.6 \\
        WTSF-ReID$^*$~\cite{yu2025wtsf} & 82.2 & 96.5 & 39.2 & 49.1 \\
        MambaPro$^\dagger$~\cite{wang2024mambapro} & 83.9 & 94.7 & 47.0 & 56.5 \\
        DeMo$^\dagger$~\cite{wang2024decoupled} & 86.2 & \textbf{97.6} & \underline{49.2} & 59.8 \\
        IDEA$^\dagger$~\cite{wang2025idea} & \underline{87.2} & 96.5 & 47.0 & \underline{62.4} \\
        \rowcolor[gray]{0.92}
        $\mathrm{\textbf{STMI}}^\dagger$ & \textbf{89.1} & \underline{97.1} & \textbf{64.8} & \textbf{76.1} \\
        \noalign{\hrule height 1pt}
    \end{tabular}
    }
    \caption{Performance on RGBNT100 and MSVR310.}
    \vspace{-1mm}
    \label{tab:multi-spectral vehicle ReID}
\end{table}

\subsection{Implementation Details}
The proposed model is implemented using the PyTorch framework and trained on an NVIDIA A800 GPU. For visual and textual encoding, we uniformly adopt the pre-trained CLIP model~\cite{radford2021learning}. The input images in the RGBNT201 dataset are resized to $256 \times 128$, while those in MSVR310 and RGBNT100 are resized to $128 \times 256$. Data augmentations include random horizontal flipping, random cropping, and random erasing~\cite{zhong2020random}. The batch size is set to 72 for the RGBNT201 dataset and 64 for the MSVR310 dataset~\cite{wang2025idea}, with 8 images sampled per identity. For the RGBNT100 dataset, a larger batch size of 128 is used, with 16 images sampled per identity. We employ the Adam optimizer to fine-tune all learnable parameters in the model, with an initial learning rate of $3.5 \times 10^{-6}$, which is gradually decayed to $3.5 \times 10^{-7}$. Additional details on prompt template, hyperparameter settings, and training efficiency are provided in the \textbf{appendix}.

\subsection{Comparison with State-of-the-Art Methods}

\textbf{Multi-Modal Person ReID.} As shown in Tab.~\ref{tab:multi-spectral person ReID}, STMI obtains \textbf{81.2\%} mAP, achieving the best performance among all compared methods. Specifically, it surpasses the previous state-of-the-art IDEA by +1.0\% in mAP. Compared with TOP-ReID (72.3\%) and EDITOR (66.5\%), STMI achieves substantial improvements of +8.9\% and +14.7\% in mAP, respectively. These results demonstrate the effectiveness of STMI in enhancing cross-modal semantic alignment and preserving token-level representation integrity.

\noindent \textbf{Multi-Modal Vehicle ReID.} As shown in Tab.~\ref{tab:multi-spectral vehicle ReID}, STMI achieves \textbf{89.1\%} mAP on the RGBNT100 dataset, outperforming strong baselines such as IDEA (87.2\%) and DeMo (86.2\%). On the more challenging MSVR310 dataset, STMI achieves \textbf{64.8\%} mAP, surpassing the previous best result by +17.8\% over IDEA (47.0\%). These results highlight the effectiveness and robustness of our method under complex conditions such as background clutter, occlusion, and modality inconsistency.

\begin{table}[t]
  \centering
  \resizebox{0.35\textwidth}{!}
  {
  \begin{tabular}{cccccc}
      \noalign{\hrule height 1pt}
      \multicolumn{1}{c}{\multirow{2}{*}{\textbf{Index}}} &\multicolumn{3}{c}{\textbf{Modules}} & \multicolumn{2}{c}{\textbf{Metrics}} \\
      \cmidrule(r){2-4} \cmidrule(r){5-6}
 & \textbf{SFM}              & \textbf{STR}                & \textbf{CHI}                   & \textbf{mAP}    & \textbf{Rank-1}   \\\hline
  A                  & \ding{53}                  & \ding{53}                  & \ding{53}                    & 70.3  & 72.1 \\
  B                  & \ding{51}                  & \ding{53}                  & \ding{53}                    & 76.1  & 78.1 \\
  \multirow{1}{*}{C} & \multirow{1}{*}{\ding{51}} & \multirow{1}{*}{\ding{51}} & \multirow{1}{*}{\ding{53}}   & 78.1  & 80.9 \\
  \rowcolor[gray]{0.92}
  \multirow{1}{*}{D} & \multirow{1}{*}{\ding{51}} & \multirow{1}{*}{\ding{51}} & \multirow{1}{*}{\ding{51}}   & \textbf{81.2} & \textbf{83.4} \\
  \noalign{\hrule height 1pt}
  \end{tabular}
  }
  \caption{Ablation study of different modules in STMI.}
  \label{tab:main_ablation}
\end{table}

\begin{figure*}[t]
  \centering
    \resizebox{0.96\textwidth}{!}
    {
  \includegraphics[width=1\linewidth]{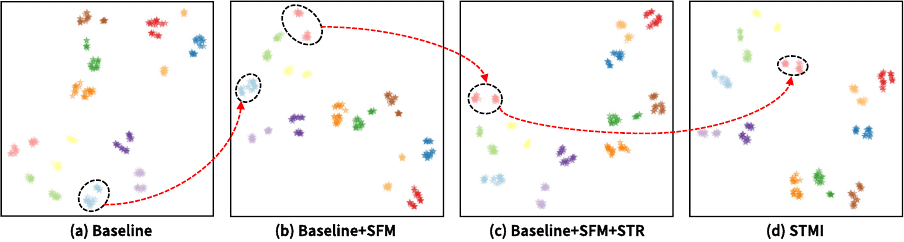}
  }
   \caption{Visualization of the feature distributions with t-SNE~\cite{van2008visualizing}.
   Different colors represent different identities.}
  \label{fig:tsne}
\end{figure*}

\begin{figure}[t]
  \centering
    \resizebox{0.475\textwidth}{!}
    {
  \includegraphics[width=1.\linewidth]{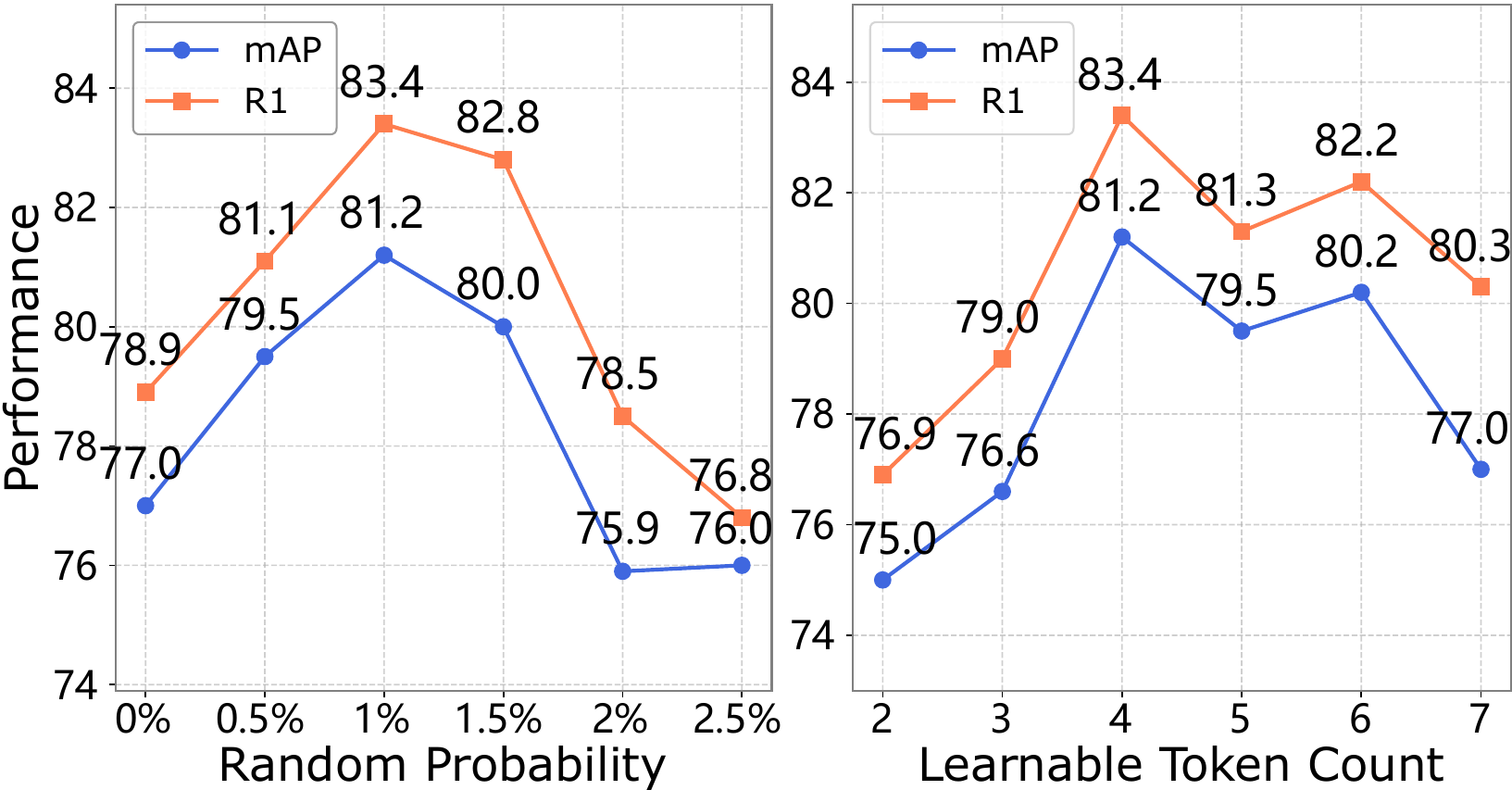}
  }
   \caption{Comparison with different hyper-parameters.}
  \label{fig:hyper}
\end{figure}

\begin{table}[t]
  \centering
  \resizebox{0.47\textwidth}{!}{
  \begin{tabular}{cccccc}
    \noalign{\hrule height 1pt}
    \textbf{Index} & \textbf{Model Variant} & \textbf{mAP} & \textbf{R-1} & \textbf{R-5} & \textbf{R-10} \\ \hline
    A & STMI (w/o CHI) & 78.1 & 80.9 & 87.9 & 90.9 \\
    B & STMI (w/ MLP, w/o CHI) & 78.0 & 82.4 & 87.3 & 90.0 \\
    C & STMI (w/ SA, w/o CHI) & 78.4 & 81.5 & 86.2 & 89.2 \\
    \rowcolor[gray]{0.92}
    D & \textbf{STMI (Full Model, w/ CHI)} & \textbf{81.2} & \textbf{83.4} & \textbf{90.2} & \textbf{91.6} \\
    \noalign{\hrule height 1pt}
  \end{tabular}
  }
  \caption{Ablation of different fusion strategies in STMI.}
      \vspace{-1mm}
  \label{tab:fusion_ablation}
\end{table}

\subsection{Ablation Studies}

We conduct ablation studies on the RGBNT201 dataset to evaluate each component of our STMI framework. The baseline adopts a three-branch vision encoder, where retrieval is based on concatenated class tokens from three modalities.

\noindent \textbf{Effects of Key Modules.} Tab.~\ref{tab:main_ablation} presents the performance of different combinations of the proposed modules. Model A serves as the baseline, achieving an mAP of 70.3\% and Rank-1 accuracy of 72.1\%. Model B incorporates the SFM module, which leverages SAM masks to enhance foreground regions and suppress background noise. This improves the mAP to 76.1\%. Model C further introduces the STR module, improving the mAP to 78.1\%. Finally, Model D integrates all three modules, including the CHI module for high-order semantic modeling via cross-modal hypergraphs, achieving the best performance with an 81.2\% mAP.

\noindent \textbf{Effects of CHI Configurations.}
To analyze the effectiveness of the CHI module, we conduct ablation experiments with different fusion strategies in the STMI framework. As shown in Tab.~\ref{tab:fusion_ablation}, Model A removes CHI entirely and directly concatenates modality-specific class tokens, achieving 78.1\% mAP. Model B replaces CHI with a MLP to fuse features across modalities, but only obtains 78.0\% mAP. Model C adopts a multi-head self-attention mechanism for cross-modal interaction, yielding 78.4\% mAP. Finally, the full model (Model D) integrates the CHI module and achieves the best performance with 81.2\% mAP. These results demonstrate that CHI effectively captures high-order semantic dependencies across modalities and provides more discriminative fusion representations compared to conventional fusion strategies.

\noindent \textbf{Effects of SFM Configurations.} 
We investigate parameter-sharing strategies in the SFM module. As shown in Tab.~\ref{tab:sfm_ablation}, sharing parameters across all layers (Model A) yields 77.2\% mAP. Modulating only the early layers (Model B) or the late layers (Model C) results in suboptimal performance. Assigning head-wise parameters (Model D) gives 77.1\% mAP. Layer-wise modulation with shared head parameters (Model E) achieves the best result (81.2\% mAP), showing the benefits of hierarchical modeling and proper parameter sharing.

\begin{table}[t]
  \centering
  \resizebox{0.475\textwidth}{!}{
  \begin{tabular}{cccccc}
    \noalign{\hrule height 1pt}
    \textbf{Index} & \textbf{SFM Setting} & \textbf{mAP} & \textbf{R-1} & \textbf{R-5} & \textbf{R-10} \\
    \hline
    A & Shared All Layers     & 77.2 & 80.9 & 86.6 & 90.3 \\
    B & Early Layers          & 73.5 & 74.3 & 83.9 & 87.9 \\
    C & Late Layers           & 76.7 & 78.9 & 87.1 & 91.3 \\
    D & Head-wise Parameters   & 77.1 & 79.9 & 86.6 & 91.0 \\
    \rowcolor[gray]{0.92}    
    E & \textbf{All Layers (Full Model)} & \textbf{81.2} & \textbf{83.4} & \textbf{90.2} & \textbf{91.6} \\
    \noalign{\hrule height 1pt}
  \end{tabular}
  }
  \caption{Ablation study on different SFM configurations.}
  \label{tab:sfm_ablation}
\end{table}
\noindent \textbf{Effects of Randomness and Token Count.}
As shown in Fig.~\ref{fig:hyper}, introducing a small amount of randomness improves generalization, while higher levels degrade performance due to excessive noise. In terms of token count, using four learnable tokens achieves the best results. Adding more tokens leads to diminishing returns and potential overfitting.

\subsection{Visualization}
\textbf{Feature Distributions.} We visualize the distribution of multi-modal features using t-SNE. As shown in Fig.~\ref{fig:tsne}, with the introduction of the SFM module, the features become more compact and identity clusters are better separated. The addition of STR further enhances intra-class compactness and inter-class separability. Finally, the full STMI model produces the clearest and most structured  distribution, demonstrating the effectiveness of each module.

\section{Conclusion}
In this work, we propose STMI, a novel framework for multi-modal object ReID that addresses the limitations of token loss and weak semantic alignment in existing methods. Specifically, we introduce the Segmentation-Guided Feature Modulation (SFM) module to enhance foreground regions and suppress background noise based on SAM-generated masks. The Semantic Token Reallocation (STR) module extracts compact and informative semantic tokens via learnable queries and cross-attention, avoiding information loss from hard filtering. Furthermore, the Cross-Modal Hypergraph Interaction (CHI) module captures high-order semantic relationships across modalities through a unified hypergraph structure. Moreover, we construct a caption generation strategy that fuses multi-modal inputs to produce reliable textual descriptions. Extensive experiments on three public multi-modal ReID benchmarks demonstrate that STMI achieves state-of-the-art performance, validating its effectiveness and generalizability.

\section{Acknowledgments}
This work was supported in part by the National Natural Science Foundation of China (No. 62362051) and the Key Scientific and Technological R\&D Program of Dalian (No. 2023YF11GX012).

\bibliography{aaai2026}

\end{document}